\documentclass[twoside,11pt]{article}

\usepackage{jmlr2e}
\usepackage{amsmath,amssymb,amsfonts}
\usepackage{algorithmic}
\usepackage{algorithm}
\usepackage{subfig}
\usepackage{float}

\usepackage{lastpage}
\jmlrheading{23}{2024}{1-\pageref{LastPage}}{4/16;2}{}{21-0000}{Bicheng Wang, Junping Wang and Yibo Xue}

\ShortHeadings{Physics-Inspired Spatial Temporal Graph Neural Networks}{Wang and Wang}
\firstpageno{1}

\begin{document}

\title{Physics-Inspired Spatial Temporal Graph Neural Networks for Predicting Industrial Chain Resilience}

\author{\name Bicheng Wang \email wangbicheng2022@ia.ac.cn \\
        \addr State Key Laboratory of Multimodal Artificial Intelligence Systems\\
       Institute of Automation, Chinese Academy of Sciences\\
       Beijing, P.R.China\\
       School of Artificial Intelligence\\
       University of Chinese Academy of Sciences\\
       Beijing, P.R.China\\
       \AND
       \name JunPing Wang \email junping.wang@ia.ac.cn \\
       \\
       \AND
       \name Yibo Xue \email xueyibo2023@ia.ac.cn \\
        }
\editor{My editor}

\maketitle

\begin{abstract}
Industrial chain plays an increasingly important role in the sustainable development of national economy. However, as a typical complex network, data-driven deep learning is still in its infancy in describing and analyzing the resilience of complex networks, and its core is the lack of a theoretical framework to describe the system dynamics. In this paper, we propose a physically informative neural symbolic approach to describe the evolutionary dynamics of complex networks for resilient prediction. The core idea is to learn the dynamics of the activity state of physical entities and integrate it into the multi-layer spatiotemporal co-evolution network, and use the physical information method to realize the joint learning of physical symbol dynamics and spatiotemporal co-evolution topology, so as to predict the industrial chain resilience. The experimental results show that the model can obtain better results and predict the elasticity of the industry chain more accurately and effectively, which has certain practical significance for the development of the industry.
\end{abstract}


\section{Introduction}

In an increasingly interconnected and complex global economy, the industrial chain becomes a vital infrastructure for safeguarding national economic security. In general, the industrial chain in \cite{2023Global} is considered to be a chaotic dynamic complex network with multiple spatiotemporal scales and nonlinear high-order interactions. It is composed of large-scale upstream and downstream user groups, products, enterprises, research and development institutions, raw material suppliers and supply chains, and forms a spatiotemporal chain correlation pattern based on specific supply and demand relations. The interaction units of enterprises in the industrial chain are often composed of three or more nodes, and there are complex nonlinear interactions between each link and subject, including dynamic nodes of different spatial scales and edge dynamics across time scales, resulting in chaos of the industrial chain. In large-scale dynamic high-order chaotic network systems similar to industrial chains, the distributed coordination control (DCC) is importance of the central nervous system, playing a crucial role in maintaining main functionalities for normal operation when the system is impacted by external events. Resilience in \cite{2024Analysis} is fundamental property of DCC system, where defined as the capacity to maintain functionality under perturbations. In light of this, all industrial physical entities operating as part of the DCC system will run in an iterative three-step resilience automate optimizing closed loop : learning resilience model from the environment, making decisions based on resilience, and responding by performing resilience regulation actions that affect the environment—either by altering operational status or via communication with other autonomous physical entities.

\begin{figure*}[h]
\centering
\subfloat[On the left is the industry chain network \cite{li2023learning}, which shows the dynamic characteristics of high-order, disorder and chaotic. Researchers aim to obtain the dynamic evolution of high-order, order and resilience through graph neural network, as shown on the right, where $C1,\cdots, Cn$ denote 
different companies]
{\includegraphics[width=.9\textwidth]{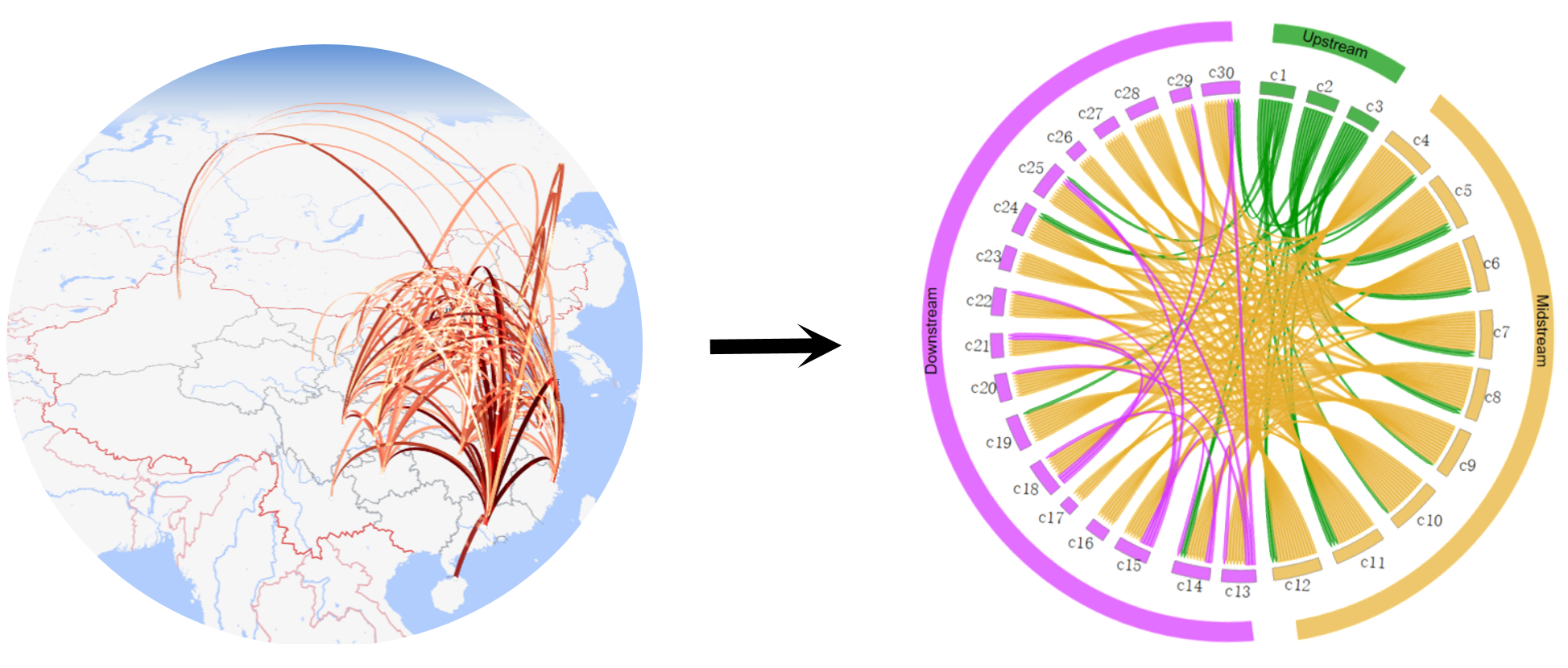}}
\centering
\hfill
\subfloat[GBB \cite{gao2016universal}: In 1D systems resilience is captured by the resilience function $x(\beta)$, which describes the state(s) of the system as a function of the tunable parameter $\beta$. The left is the bifurcating resilience function. The system exhibits a single stable fixed point for $\beta>\beta_c$ (blue) and two (or more) stable fixed points, a desired (blue) and an undesired (red) for $\beta<\beta_c$. The middle is resilience function with a first-order transition from the desired (blue) state to the undesired (red) state. The right is resilience function with a stable solution for $\beta<\beta_c$ and no solution above $\beta_c$, resulting in an uncontrolled divergent or chaotic behavior.]
{\includegraphics[width=.48\columnwidth]{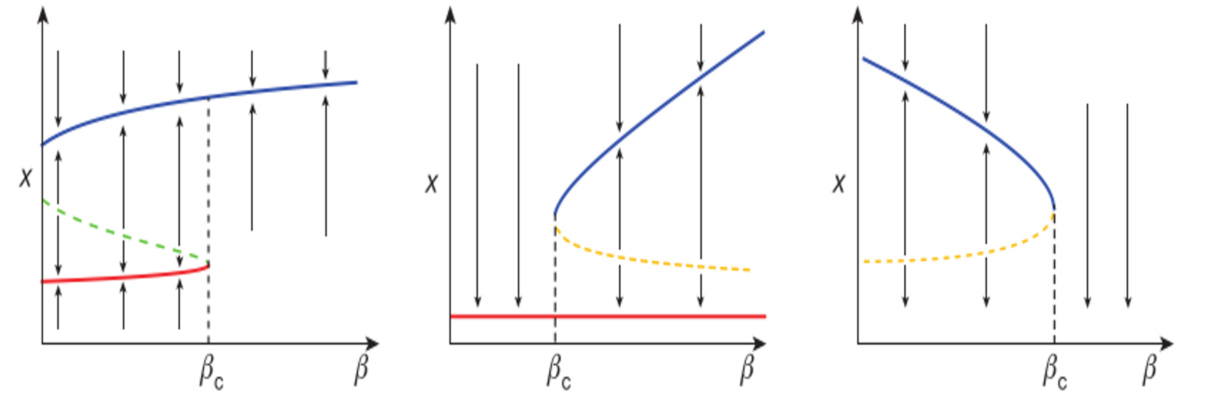}}\label{gbb}
\centering
\hfill
\subfloat[\cite{liu2024deep} simulate the node activities using the corresponding dynamics on mutualistic dynamics and gene regulatory dynamics network, and employ kernel density estimation (KDE) plot to visualize the distribution of its stable states. The ground truth simulations show Network (II) and (III) are resilient because they have a unique, non-trivial stable states (〈x〉$>$ 0), while Network (I) and (IV) are non-resilient for having more than 1 stable states or only 1 trivial stable state (〈x〉 =0), exhibiting chaotic behavior, that is, the interaction dynamics do not settle down to periodic or stationary points, but continue to oscillate in an irregular manner that is aperiodic.]
{\includegraphics[width=.5\columnwidth]{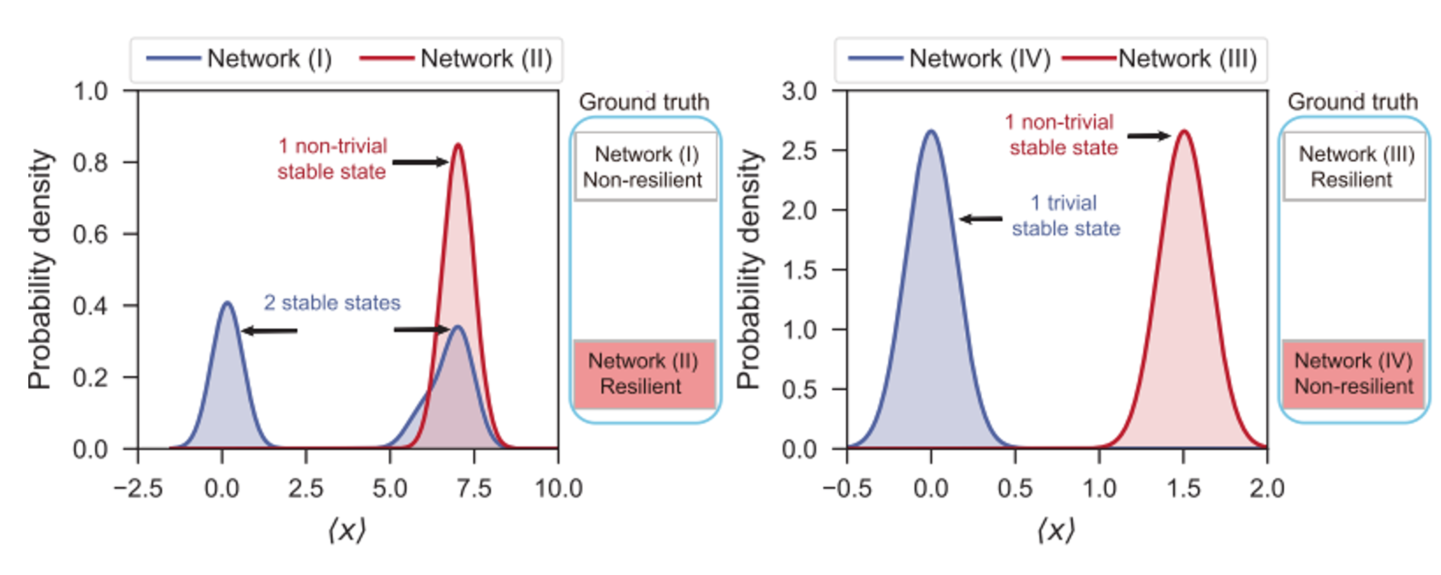}}
\centering
\caption{Advanced Research on Supply Network Network Resilience . }
\label{fig1}
\end{figure*}

The prediction of industrial chain resilience is the key to realize automatic and accurate optimization of DCC. Current resilience prediction research shows two trends. On the one hand, it focuses on the empirical study of industry chain elasticity. \cite{gao2016universal}, \cite{zhang2022estimating}, \cite{artime2024robustness} attempted to provide analytical estimates for the resilience of n-dimensional systems with complex interactions between components by reducing them to tractable 1-D systems based on mean-field theory, spectral graph theory, etc. At present, the most widely used is the Gao-Barzel-Barabasi (GBB) framework \cite{gao2016universal}, as shown in Figure. \ref{fig1}, GBB reduces the network system to a one-dimensional nonlinear equation $\frac{dx}{dt}=f(\beta,x)$ and calculates a single elastic parameter $\beta\in \mathbb{R}$. A system is considered resilient only if its resilience parameter exceeds a certain critical threshold, that is, $\beta>\beta_c$. However, the accurate estimation of $\beta$ and $\beta_c$ relies on a strong assumption that the node activity dynamics can be described by linear equations and that the interconnection degrees of nodes are almost independent of each other. Despite extensive research on physical system resilience and related topics such as alternative steady states, state transitions, and feedback loops, there are far more theoretical studies of resilience than empirical ones, and they often make strong assumptions about network topology and dynamics for analytical feasibility. For example, \cite{2001Catastrophic} compared to the plethora of mathematical models demonstrating the existence of alternative stable states, there are only a few empirical studies, especially experimental studies conducted in the real world rather than in laboratories. The reason for this is that most real networks are too complex and interconnected to permit real-world experimentation. There are always gaps between theoretical models and natural systems; for example, the universal dimension-reduction method for predicting the resilience of complex networks is based on the assumption that no negative interactions occur in networks.

On the other hand, focuses on uncovering network dynamics from data on system evolution by deep graph learning, see \cite{2024Analysis}. \cite{li2024predicting}, \cite{liu2024tdnetgen} use the deep learning method to fit the dynamic evolution of the network from the data, which often requires a complex network and high quality data. These studies assume that all nodes in a given industrial network follow the same dynamical pattern, whereas nodes in a real network may exhibit very different dynamic behavior. For example, in a protein interaction network, two proteins may be connected due to their simple physical chain binding/separation or because they participate in the same metabolic reaction that produces other compounds. the resilience dynamic models presented in these studies are all ordinary differential equations, and time is the argument. Other variables, such as spatial distance, could be vital driving forces in dynamic changes. Hence, a more general model represented by partial differential equations may be needed. each component’s time scale may vary over a vast range from seconds to months, a fact that brings significant challenges to dynamic modeling. A feasible way in which to consider all these complexities is to combine resilience of the real-world system with data-driven dynamic model discovery methods.   

Most approaches of industrial chain resilience prediction rely on predefined equations for physical entities activity dynamics and simplifying assumptions on supply chain network topology. For example, the industrial chain is essentially a dynamic, spatiotemporal co-evolution system. Large-scale physical entities states evolution of complex networks is driven by underlying nonlinear dynamics. Specifically, through the function of each node represented by its state value, a resilient network can recover from outages (at its nodes) and dynamically evolve to a stable phase where all nodes operate at high activity levels. It is increasingly clear that networks in the real world are not isolated but are interdependent with one another. How to accurately identify and construct spatiotemporal co-evolutionary dynamics models has become a key bottleneck restricting the prediction of industrial chain resilience, due to their large interconnected nodes, unstable supply chain status, less available information, and many other objective reasons in industrial chain, limiting their applicability to real-world systems.

Based on this observation, this paper proposes physics-informed neural-symbolic learning method integrating co-evolutionary dynamics non-linear differential equation into multi-layer spatiotemporal dynamic network (MSTDN), where is to infer resilience directly from observational data. Specifically, this proposed scheme constructs a parametrized dynamics model for each physical entities using ordinary differential equations. Then, it trains multi-layer spatiotemporal dynamic network model of industrial chain from large-scale high-dimensional sensory, in which it simultaneously trains three models: spatial and temporal layers, dynamic weights and recursive connections, generalizing optimal of policy across multiple known and unknown tasks. All the related parametrized state for each node are integrated into the multi-layer spatiotemporal dynamic network, which is learning global spatiotemporal co-evolution dynamic by physics-informed neural-symbolic, transferring control policy at every layer in a dynamic multi-layer spatiotemporal network. so as to predict the resilience of the industrial chain network. The main contributions in this paper are as follows:

\begin{itemize}
    \item The dynamics model of large-scale physical entities activity states is developed using the non-linear ordinary differential equations. This model employs stacked transformer encoder layers to generate representations for the governing equations of physical entities activity dynamics by modeling the complex correlations among activities. Dynamic models are able to achieve a desired trade-off between accuracy and efficiency for dealing with varying computational budgets on the fly. Therefore, they are more adaptable to different heterogeneous system and changing environments, compared to static models with a fixed computational cost.
    \item The multi-layer spatiotemporal dynamic network (MSTDN) is automatically constructing from large-sacle industrial chain co-evolutionary data. We design a spatiotemporal topology encoder that uses spatiotemporal neural network to model the large-scale physical entities co-evolutionary network topology from the input adjacency matrix with a message-passing mechanism that recursively aggregates features from neighboring nodes. It can flexibility generate discriminating topological representations for large-scale physical entities co-evolution resilience.
    \item Considering interpretability and accuracy of industrial chain resilience prediction, we will integrate the dynamics model of physical entities activity states into multi-layer spatiotemporal dynamic network, where defines physics-informed neural-symbolic. The physics-informed neural-symbolic is possible to analyze which components of a deep model are activated when processing an input instance, and to observe which parts of the input are accountable for certain predictions. These properties not only may shed light on interpreting the prediction process of MSTDNs, but also may enhancing prediction accuracy in spatiotemporal co-evolutionary networks.
\end{itemize}

The remainder of the paper is organized as follows:
Section 2 introduces related works covering industrial chain resilience prediction. Section 3 describes the proposed physics-informed neural-symbolic model in detail, give physics-informed neural-symbolic algorithm to solve spatiotemporal co-evolutionary dynamics problems with large-scale state
spaces, analyze its computational complexity and convergence,
and provide some discussions. Experimental analysis and completion results are shown in Section 4 to verify our method.
Finally, we conclude the proposed method in Section 5.

\section{Related Works}

The dynamic processes of real-world complex systems, characterized by nonlinearity and multi-scale, are often abstracted into complex network models, thus illustrating the interactions between nodes. With the development of deep learning techniques such as graph neural networks, data-driven modeling of complex network dynamics has attracted a lot of attention. \cite{murphy2021deep} propose a GNN architecture that can accurately model disease spread on the network under minimal assumptions on the dynamics. \cite{zang2020neural} was the first to combine neural ODEs and GNNS to model the continuous-time dynamics of complex networks. \cite{huang2023generalizing} successfully modeled dynamic topology and cross-environment network dynamics by introducing edge dynamic Odes and environment encoders, respectively. Network resilience is an important indicator to describe the dynamics of complex networks and dynamic complex systems. \cite{laurence2019spectral} used spectral graph theory to reduce the dimensionality of the dominant eigenvalues and eigenvectors of the adjacency matrix. \cite{morone2019k} developed a resilience prediction method by quantifying the k-core structure in the network. The data-driven based resilience prediction method proposed by \cite{liu2024tdnetgen} uses the inherent joint distribution existing in unlabeled network data to facilitate the learning process of the resilience predictor by elucidating the relationship between network topology and dynamics.

Although some effective results have been achieved, most of them are data-driven, which are difficult to effectively describe complex networks, lack interpretability and can only predict discrete data. In our work, we introduce symbolic logic as a rule for the dynamic evolution of networks, using ODE solvers that can describe the dynamic changes of networks in continuous time.
It has gradually become a new consensus to let neural networks play their advantages as powerful perception models and use complex reasoning parts to deal with symbolic systems, that is, to utilize neuro-symbolic learning \cite{yu2023survey}\cite{li2020closed}. At present, there are three main ways of neural symbolic learning: learning aided reasoning, reasoning aided learning and learning reasoning. Learning-aided reasoning methods mainly use symbolic reasoning to solve the problems in machine reasoning, and introduce neural networks to assist the solution. \cite{qu2019probabilistic} combined GCN with Markov logic network to solve the task of knowledge graph reasoning. Reasoning assisted learning mainly uses symbolic methods such as logical reasoning and knowledge graph to constrain the learning results of neural networks. \cite{xie2019embedding} integrated propositional logic into the relationship detection model, and proposed a Logic embedding network with semantic regularization (LENSR) to enhance the relationship detection ability of the deep model. \cite{yu2022probabilistic} built a global dependency graph for all logical formulas, integrated symbolic knowledge into the deep learning model, and proposed a two-layer probabilistic graphical reasoning framework to improve the performance and interpretability of the visual relation detection model. In physics, neural symbolic learning extracts models of objects and relations in an unsupervised manner \cite{van2018relational}. \cite{greydanus2019hamiltonian}] couple GNNS with differentiable ODE solvers and model their interactions as dynamic graphs for learning Hamiltonian dynamics of physical systems. In the learning-inference approach, the interaction between neural networks and symbolic reasoning is bidirectional. Based on ProbLog, \cite{manhaeve2018deepproblog} introduced the concept of neural fact and neural annotation parsing, and proposed a model that seamlessly integrates probability, logic and deep learning, namely DeepProbLog. DeepProbLog is a pioneering framework that enables end-to-end training of neural networks and logical inference. Specifically, deep learning is responsible for mapping the input unstructured data into a distribution of categories, called neural predicates, which are the interfaces that connect deep learning models and logical reasoning. The logical inference module describes the problem via ProbLog and models it as a directed graph using arithmetic loops. Where the root node is the query, and the leaf nodes are neural predicates and other (non-neural network outputs) probabilistic facts.

\section{Problem Formulation}

\begin{figure*}[h]
\centering
\subfloat[A description of industrial chain topology resilience.In the upstream, midstream and downstream of the industrial chain, there are enough associated enterprises. When some enterprises in the industry chain change, a resilient topology will maintain the stability of the industry chain, while a non-resilient industry chain may cause a certain part to collapse.]
{\includegraphics[width=.9\textwidth]{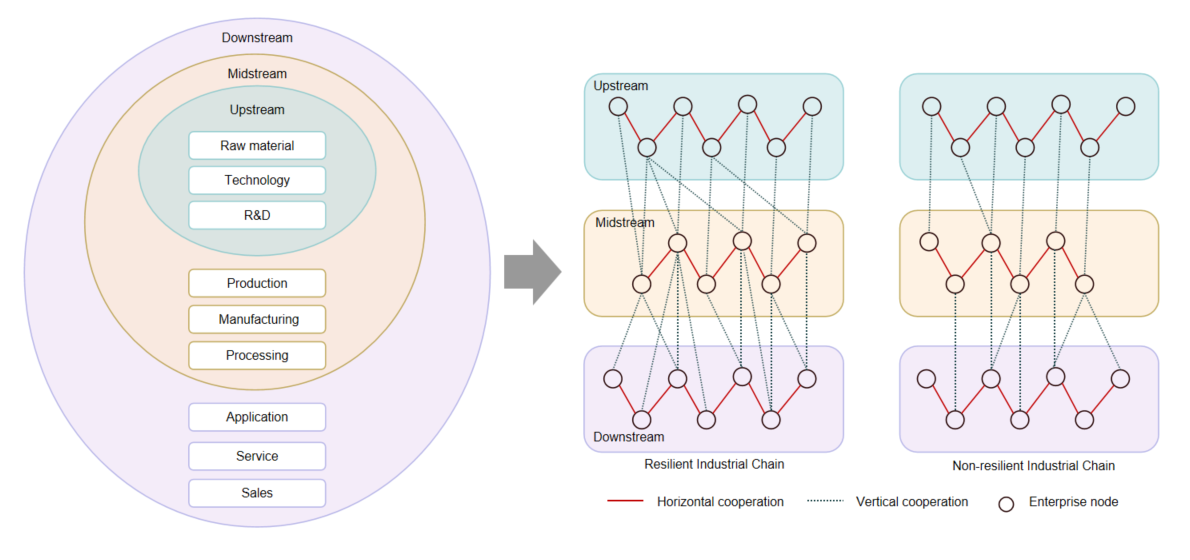}}
\centering
\hfill
\subfloat[A description of industrial chain node state resilience.For the resilient network, after receiving the disturbance, it will recover to the desired stable state for a period of time, while the non-resilient network will not recover.]{\includegraphics[width=.9\textwidth]{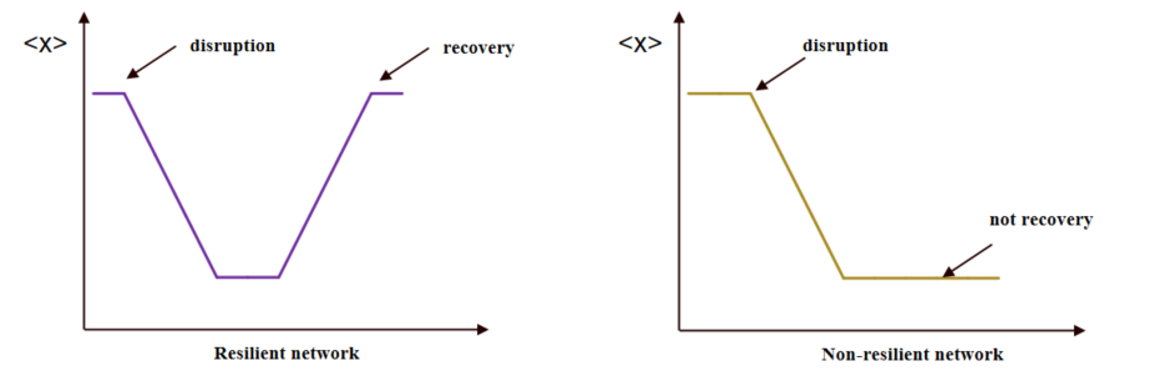}}
\centering
\caption{A description the the industrial chain resilience.}
\label{re}
\end{figure*}

Network resilience is an important indicator to describe the dynamics of complex networks and dynamic complex systems. Generally, it refers to the ability of the network to take the initiative to deal with destructive events, maintain dynamic balance, and restore normal operation when some links have problems. Network resilience clarifies that resilient systems are characterized by always tending to desirable, nontrivial stable equilibria after perturbations.
Generally, network resilience is described by two aspects: on the one hand, whether the industry chain will generate new links to maintain the normal operation of the industry chain after receiving a disturbance. As shown in Figure., we use different colors to represent the relationship between industry chains in different years. When a chain in the industry chain changes, new chains will be generated over time to maintain the normal operation of the industry chain. On the other hand is the change of node state. Formally, given a complex network $\mathcal{G}=(\mathbf{\mathcal{V}},\mathbf{\mathcal{E}},\mathbf{A})$, where $\mathbf{\mathcal{V}}$ denotes the set of its nodes, $\mathcal{E}$ denotes the set of links and $\mathbf{A}$ denotes the adjacency matrix. The state of node $i$ can be expressed as $\mathbf{u}_i$, and is usually governed by the following nonlinear ordinary differential equation (ODE) as the node state dynamics:
\begin{eqnarray}\label{dudi}
\frac{d\mathbf{u}_i}{dt}=F(\mathbf{u}_i)+\sum_{j=1}^NA_{ij}G(\mathbf{u}_i,\mathbf{u}_j)
\end{eqnarray}
where $F(\mathbf{u}_i)$ represents the self-dynamics of nodes and $G(\mathbf{u}_i,\mathbf{u}_j)$ denotes interaction dynamics. The complex network $\mathcal{G}$ is considered resilient if it consistently converges to only the desired nodal state equilibrium as time $t$ approaches infinity, irrespective of any perturbation and varying initial conditions with the exception of its fixed points. The complex network resilience curve is shown in Figure. \ref{re}.

\section{Methodology}

\begin{figure*}[t!]
    \centering
    \includegraphics[width=\textwidth]{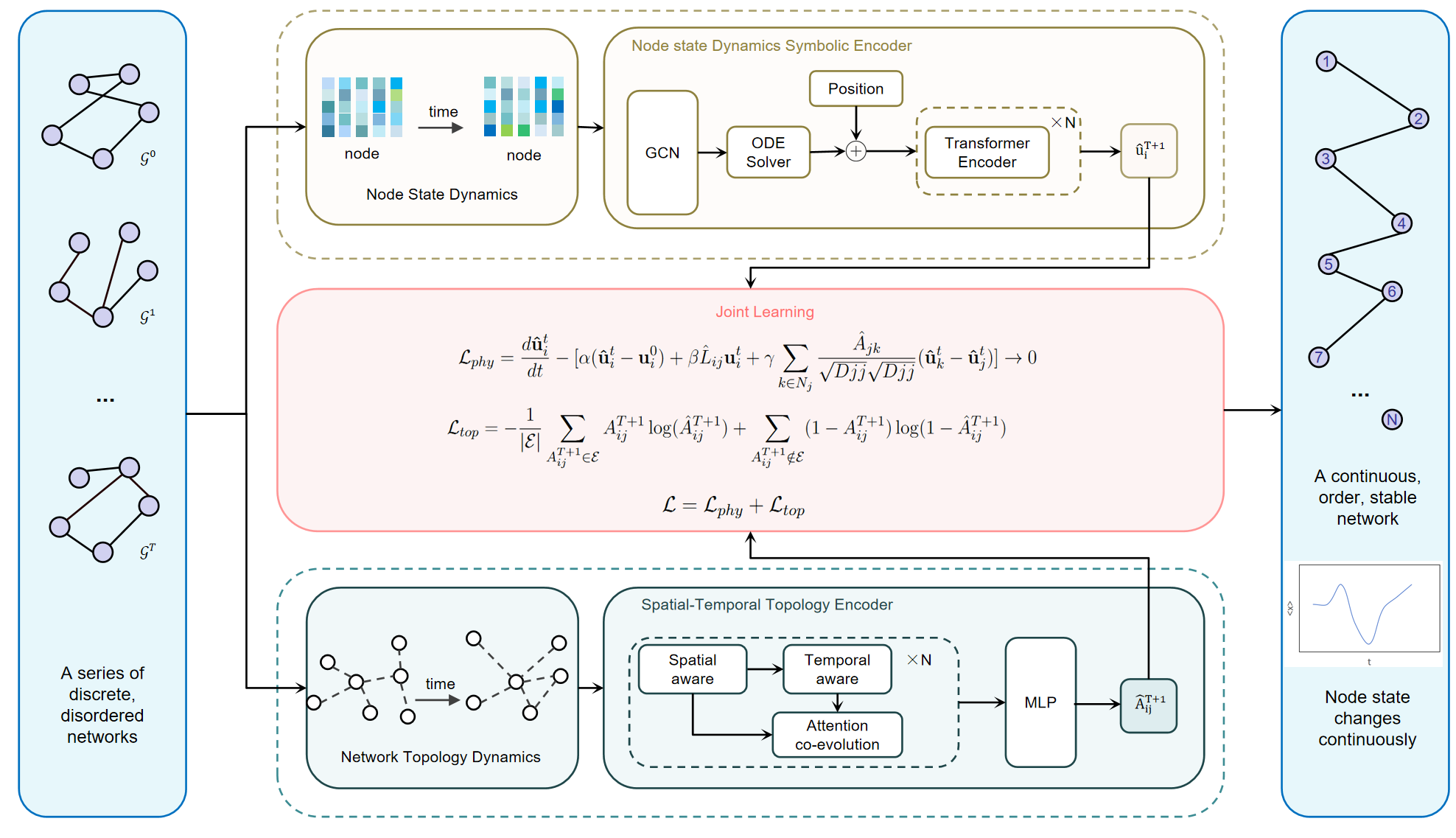}
    \caption{The framework of NSL-Net. a. We use two-layer GCN to extract node features and multi-layer Transformer Encoder to generate the representation of entity activity dynamic governing equations. b. We construct a spatiotemporal topology encoder to model a large-scale physical entity co-evolutionary network topology with input adjacency matrix using spatiotemporal graph neural network, and use attention mechanism to aggregate spatiotemporal information to generate topological representation. c. We combine the dynamic model of physical activity state with the multi-layer space-time dynamic network to learn, and optimize the model by using the constructed neural-symbolic differential equation with physical knowledge and cross entropy as loss function.Finally, we get a continuous, order, stable network whose node states change continuously.}
    \label{framewprk}
\end{figure*}

\subsection{Modeling physical entity symbolic activity dynamics}\label{preli}

As mentioned in Section \ref{preli}, node state dynamics are usually described by a nonlinear ordinary differential equation as shown in Equation \ref{dudi}. The purpose of this section is to analy and obtain differential equations for the dynamics of graph networks by means of diffusion equations from physics.

We input the node state trajectory $\mathbf{U} \in \mathbb{R}^{M\times N\times T}$, where $M, N, T$ represents the node state dimension, node number, and time dimension respectively. First, we use a two-layer GCN to extract the node features of each timestamp, which can be expressed as:
\begin{eqnarray}
    \mathbf{e}^t_i \in \mathbb{R}^{d_e}&=& GCN(\mathbf{u}^t)
\end{eqnarray}

Then, the node features are computed using a numerical ODE solver such as Neural ODE. Compared with time series prediction models based on deep learning, the ODE solver has the advantage that it only needs to know the initial state and the known equation to obtain the solution at any desired time, which is beneficial to solve data with inconsistent time intervals. The general form of an ODE solver is as follows:

\begin{eqnarray}
    \mathbf{e}_i^T = ODESolver(g,[\mathbf{e}_1^0,\mathbf{e}_2^0,\cdots,\mathbf{e}_N^0],t=T)=\mathbf{e}_i^0+\int_{t=0}^{T} g(\mathbf{e}_1^t,\mathbf{e}_2^t,\cdots,\mathbf{e}_N^t)dt
\end{eqnarray}

We also add positional embedding to the input to mark the relative positions, i.e., $\mathbf{e}^t_i = \mathbf{e}^t_i +PE$. At each time step $t$ of the total d time steps, positional embeddings are formulated as $PE_{(t,2n)}= =sin(t/10000^{2n/d_e}),PE_{(t,2n+1)} =cos(t/10000^{2n/d_e})$, where the second element in the tuple of the subscript $(2n$ and $ 2n+1)$ corresponds to the index of elements in positional embeddings.

Each Transformer Encoder consist of a self-attention sub-layer and a fully connected feed-forward network (FFN) sub-layer. A residual connection is established between the two sub-layers, followed by layer normalization.

The self-attention sub-layer consist of multiple attention heads, which allows us to integrate information from different representation subspaces simultaneously, while considering different positional contexts. For the attention $head_j$, the result embedding can be obtained from the query matrix $\mathbf{Q}_j$ and the key, value $\mathbf{K}_j, \mathbf{V}_j$, which is formulated as:
\begin{eqnarray}
    head_j &=& softmax(\frac{\mathbf{Q}_j\mathbf{K}^T_j}{\sqrt{d_k}}\mathbf{Q}_j)\\
    \mathbf{Q}_j &=& \mathbf{e}^t_i\mathbf{W}^Q_j\\ \mathbf{K}_j &=& \mathbf{e}^t_i\mathbf{W}^K_j\\ \mathbf{V}_j &=& \mathbf{e}^t_i\mathbf{W}^V_j
\end{eqnarray}
where $\mathbf{W}^Q_j, \mathbf{W}^K_j, \mathbf{W}^V_j$ are trainable parameters.

By fusing the information from h attention heads, we can get a multi-encoded representation, which is also used as the output of the self-attention sub-layer, so the final output of the self-attention sub-layer can be formulated as:
\begin{eqnarray}
    att(\mathbf{e}^t_i) &=& [head_1, head_2, \dots, head_h]\mathbf{W}_O\\
    \mathbf{e}'^t &=& LayerNorm(\mathbf{e}^t_i+att(\mathbf{e}^t_i)
\end{eqnarray}
where $\mathbf{W}_O $ is trainable parameter.

The FFN sub-layer consist of two linear transform layers which the ReLU activation function is applied. Therefore, the final output of the fully connected feed-forward sub-layer can be formulated as:
\begin{eqnarray}
    \mathbf{\hat{e}}^t &=& LayerNorm(\mathbf{e'}^t+FF(\mathbf{e}^t_i))\\
    FF(\mathbf{e}^t_i) &=& RELU(\mathbf{e}^t_i\mathbf{W}_{f1}+\mathbf{b}_{f1})\mathbf{W}_{f2}+\mathbf{b}_{f2}
\end{eqnarray}

We use a multi-layer Transformer Encoder to act as a dynamic encoder and finally get $\mathbf{u}^{T+1}_i$ as the predicted future node state.

We then consider modeling differential equations that describe the dynamics of the physical entity's activity. We consider the reaction-diffusion equation. The RD equation has wide applications in pattern formation analysis, such as population dynamics, chemical reactions, etc. In general, the RD system can be described by the following control equation:
\begin{eqnarray}
\mathbf{u}_t &= &\mathbf{D} \Delta\mathbf{u}+\mathbf{R}(\mathbf{u} ) 
\end{eqnarray}
where $\mathbf{u}\in \mathbb{R}^n$ is the vector of concentration variables, the subscript $t$ denotes time derivative, $n$ represents the system dimension, $\mathbf{D}\in \mathbb{R}^{n\times n}$ is the diagonal diffusion coefficient matrix, $\Delta$ is the Laplacian operator and $\mathbf{R}(\mathbf{u})$ is the reaction vector that represents the interactions among components of $\mathbf{u}$.

We start with the Laplacian operator of calculus. The Laplacian operator for a multivariate function $f(x_1,x_2\dots,x_n)$ is the sum of the non-mixed second-order partial derivatives of all independent variables: 
\begin{eqnarray}
    \Delta f&=&\sum_{i=1}^{n}\frac{\partial ^2f}{\partial x_{i}^2} 
\end{eqnarray}

It can be interpreted as the sum of the rate of change of f at $(x_1, x_2,\dots ,x_n)$ in all directions.

Then, we consider a graph $\mathcal{G}$. Its adjacency matrix $\mathbf{A}$ is a $n\times n$ matrix, and the matrix element $a_{ij}$ represents the weight of edge $(i,j)$. If there is no edge connection between two vertices, the corresponding element in the adjacency matrix is 0. And a degree matrix $\mathbf{D}$ that is a diagonal matrix can be expressed as $D_{ii}=\sum_{j=1}^{n}A_{ij}$. 

Generalized the Laplacian operator to the graph. If the values at the vertices of the graph are regarded as function values, the Laplacian operator at the node is $\Delta f_i = \sum_{j\in N_i}A_{ij}(f_i-f_j)=D_if_i-\mathbf{A}_i\mathbf{f}$. For all vertices of the graph, there is: 
\begin{eqnarray}
\Delta f &=& \begin{bmatrix}
\Delta f_1 \\ \cdots \\ \Delta f_n
\end{bmatrix} 
=
\begin{bmatrix}
 d_1f_1-\mathbf{A}_i\mathbf{f}  \\\dots \\d_nf_n-\mathbf{A}_n\mathbf{f}
\end{bmatrix}\\&=&
(\mathbf{D-\mathbf{A} } )\mathbf{f} 
\end{eqnarray}

Fortunately, in graph theory, the Laplacian matrix can be obtained as:
\begin{eqnarray}
    \mathbf{L}&=&\mathbf{D}-\mathbf{A}
\end{eqnarray}

The Laplacian matrix is a real symmetric positive semi-definite matrix with eigenvalues greater than or equal to 0. An eigenvalue of 0 represents nodes that are not connected to each other, and the number of eigenvalues of 0 represents the number of nodes that are not connected to each other.

In fact, the value of the Laplacian matrix at a certain position $(L_{ij})$ is the second-order non-mixed partial derivative of the signal along the $ij$ edge on the graph that is, the Laplacian matrix is the Laplacian operator on the graph. And the $i$-th row in the Laplacian matrix actually reflects the cumulative gain generated by the $i$-th node when it perturbs all other nodes.

Similar to the RD equation, the differential equation of node states in the network can be expressed as follows:
\begin{eqnarray}
    \frac{d\mathbf{u}_i^t}{dt}&=& L_{ij}\mathbf{u}_i^t
\end{eqnarray}

The diffusion equation represents the dynamics of the node state $\mathbf{u}_i^t$ with respect to $t$, which depends on the difference between nearby nodes, meaning that the greater the difference between a node and its neighbors, the faster it will change. However, the speed of change cannot depend solely on the difference in representation between node $i$ and its neighbors, otherwise it will lead to the over-smoothing problem, that is, nodes cannot be distinguished as the diffusion process proceeds. As in Equation \ref{dudi}, the auto-dynamic term of the node should also be added, which is denoted as $\mathbf{u}_i^t-\mathbf{u}_i^0$.

The above analysis is only for the first-order diffusion process, that is, the diffusion occurs only between the node and its 1-hop neighbor nodes. In order to utilize 2-hop neighbor nodes, a regularization term is added to the equation $\sum_{k\in N_j}\frac{A_{jk}}{\sqrt{Djj}\sqrt{Djj}}(\mathbf{u}_k^t-\mathbf{u}_j^t)$. \cite{li2024generalized}

In addition, in order to improve the generalization of the equation, we add hyperparameters $\alpha, \beta, \gamma$ to the equation and adjust different hyperparameters for different data sets. Therefore, we finally obtain the differential equation of node dynamics as follows:
\begin{eqnarray}
    \frac{d\mathbf{u}_i^t}{dt}&=& \alpha(\mathbf{u}_i^t-\mathbf{u}_i^0)+\beta L_{ij}\mathbf{u}_i^t +\gamma \sum_{k\in N_j}\frac{A_{jk}}{\sqrt{Djj}\sqrt{Djj}}(\mathbf{u}_k^t-\mathbf{u}_j^t)
\end{eqnarray}

\subsection{Spatiotemporal relational co-evolution topology learning}\label{MSTDN}

The top of the Figure. \ref{framewprk} shows the overall structure of the topological structure prediction module. First, given the temporal graph at time $t$, where the node state is denoted by $\mathbf{u}_i^t$, the set of its neighbor nodes is denoted by $N_i^t$, and the link to neighbor node $j$ is denoted by $e_{i,j}^t$, the spatial aware is defined as:
\begin{eqnarray}
    \mathbf{z}_i^t&=&\phi(\mathbf{u}_i^t,\{(\mathbf{u}_j^t,e_{i,j}^t);\Theta_{z}^t\})
\end{eqnarray}\label{zit}
where $\Theta_{z}^t$ is the learnable parameter. Considering that different neighbors have different influences on the target node, the linear attention operator is adopted and \ref{zit} is modified as follows:
\begin{eqnarray}
    a_{ij} &=& \frac{\mathrm{exp}(\mathbf{v}_{z}^t\sigma(\mathbf{W}_{z_1}^t[\mathbf{u}_i^t,\mathbf{u}_j^t,e_{i,j}^t]))}{\sum_{j' \in N_i^t}\mathrm{exp}(\mathbf{v}_{z}^t\sigma(\mathbf{W}_{z_1}^t[\mathbf{u}_i^t,\mathbf{u}_{j'}^t,e_{i,j'}^t]))}\\
    \mathbf{z'}_i^t &=& \sigma(\mathbf{W}_{z_2}^t \sum_{j\in N_i^t} a_{ij}[\mathbf{u}_j^t,e_{i,j}^t])\\
    \mathbf{z}_i^t &=& \sigma(\mathbf{W}_{z_3}^t[\mathbf{u}_i^t,\mathbf{z'}_i^t])
\end{eqnarray}
where $\sigma$ is a nonlinear activation function (sigmoid function in our implementation), $[\cdot,\cdot]$ represents the concatenation of vectors, $\mathbf{W}_{z_1}^t, \mathbf{W}_{z_2}^t, \mathbf{W}_{z_3}^t$ and $\mathbf{v}_{z}^t$ are the learnable parameters of the spatial aware at time $t$. By iteratively stacking L times, the final spatial information is able to capture the topological information and attribute information of the L-hop neighborhood. To simplify the presentation, we still use $\mathbf{z}_i^t$ to denote the final spatial information of node i at time t.

For node $i$, given the spatial information of its T temporal graphs $\{\mathbf{z}_i^t\}_{t=0}^T$, the temporal aware takes the form of LSTM, which can be defined as follows:
\begin{eqnarray}
    \mathbf{h}_i^t &=& LSTM(\{\mathbf{z}_i^t\}_{t=0}^T;\Theta_{h})
\end{eqnarray}
where $\Theta_{tem}$ is the learnable parameter of the temporal aware.

Then given the aggregation of temporal and spatial information for node $i$, defined as $\mathbf{E}=\{\mathbf{z}_i^t\}_{t=0}^T \cup \{\mathbf{h}_i^t\}_{t=0}^T$. Using the attention operator, we can generate aggregated information:
\begin{eqnarray}
    \mathbf{q}_i &=& \sigma(\mathbf{v}_{e_1}^T\sum_{\mathbf{e}\in \mathbf{E}}\delta_e \mathbf{e})\\
    \delta_e &=& \frac{\mathrm{exp}(\mathbf{v}_{e_2}^T\sigma(\mathbf{e}))}{\sum_{\mathbf{e'}\in \mathbf{E}}\mathrm{exp}(\mathbf{v}_{e_2}^T\sigma(\mathbf{e'}))}
\end{eqnarray}
Where $\delta_e$ denotes the normalized attention score. Thus, the final information $\mathbf{q}_i$ is able to adaptively capture spatiotemporal information from the neighborhood of node $i$ of T temporal graphs.

Finally, the predicted topology is obtained by MLP consisting of two fully connected layers:
\begin{eqnarray}
    \hat{A}_{ij}^{T+1}&=&MLP(\mathbf{q}_i)
\end{eqnarray}

\subsection{Joint learning resilience of physical symbolic dynamics and spatiotemporal co-evolution topology}\label{Integrating}

We combined the physical symbol dynamic u with the spatiotemporal co-evolutionary topology A to predict network toughness. We take the constructed node dynamic differential equation as a loss function, which is formalized as:
\begin{eqnarray}
    \mathcal{L}_{phy}&=&\frac{d\mathbf{\hat{u}}_i^t}{dt}-[ \alpha(\mathbf{\hat{u}}_i^t-\mathbf{u}_i^0)+\beta \hat{L}_{ij}\mathbf{u}_i^t +\gamma \sum_{k\in N_j}\frac{\hat{A}_{jk}}{\sqrt{Djj}\sqrt{Djj}}(\mathbf{\hat{u}}_k^t-\mathbf{\hat{u}}_j^t)] \to 0
\end{eqnarray}

In addition, the topology prediction task is usually defined as a binary classification problem, and the cross entropy is set as a loss function as follows:
\begin{eqnarray}
    \mathcal{L}_{top} &=& -\frac{1}{|\mathcal{E}|}\sum_{A_{ij}^{T+1}\in \mathcal{E}}A_{ij}^{T+1}\log(\hat{A}_{ij}^{T+1})+\sum_{A_{ij}^{T+1}\notin \mathcal{E}}(1-A_{ij}^{T+1})\log(1-\hat{A}_{ij}^{T+1})
\end{eqnarray}

Therefore, our proposed joint learning loss function is:
\begin{eqnarray}
    \mathcal{L}&=& \mathcal{L}_{phy}+\mathcal{L}_{top}
\end{eqnarray}

The overall algorithmic flow of our method is shown in Algorithm. \ref{alg}.

\begin{algorithm}[h]
    \caption{Algorithm of NSL-Net}
    \label{alg}

    \begin{algorithmic}[1]\label{alg}
        \REQUIRE  a spatiotemporal complex networks $\mathcal{G} $, the feature matrix $\mathbf{U}$, the topology $\mathbf{A}$
        \ENSURE  $\hat{A}_{ij}^{T+1}$, $\mathbf{\hat{u}}^{T+1}$

       \FOR{$t \in [0,T]$}
    
            \STATE $\mathbf{e}^t_i \in \mathbb{R}^{d_e} = GCN(\mathbf{u}^t)$;\\   
            //Extract node dynamic features
            \STATE $\mathbf{u}^{T+1}_i = Transformer Encoder(\mathbf{e}^t_i)$;\\
            //Obtain the future node state

        \ENDFOR
        
        \FOR{$t \in [0,T]$}

            \STATE $\mathbf{z}_i^t=\phi(\mathbf{u}_i^t,\{(\mathbf{u}_j^t,e_{i,j}^t);\Theta_{z}^t\})$;\\
            // Capture spatial association information
            \STATE $\mathbf{h}_i^t = LSTM(\{\mathbf{z}_i^t\}_{t=0}^T;\Theta_{h})$;\\
            // Capture temporal association information
            \STATE $\mathbf{E}=\{\mathbf{z}_i^t\}_{t=0}^T \cup \{\mathbf{h}_i^t\}_{t=0}^T, \mathbf{q}_i =\sigma(\mathbf{v}_{e_1}^T\sum_{\mathbf{e}\in \mathbf{E}}\delta_e \mathbf{e})$;\\
            //Aggregate spatiotemporal information
            
        \ENDFOR
         
        \STATE $\hat{A}_{ij}^{T+1} \gets MLP(\mathbf{q}_i)$; \\
        //Obtain the network topology by neural network learning

        \STATE $\mathcal{L}= \mathcal{L}_{phy}+\mathcal{L}_{top}$;\\
        \STATE Minimize $\mathcal{L}$ to update network parameters;\\
        //Joint Learning resilience

    \end{algorithmic}
\end{algorithm}

\section{Experiments}

In this section, we construct complex networks using industrial chain data obtained from the real world and use it to evaluate the effectiveness of our proposed model.

\subsection{Experiments setup}

An industrial chain network is a complex system that describes the interdependence of an enterprise, industry or market. It is usually composed of multiple nodes, which represent the entities participating in the industrial chain. We obtained the operational data of enterprises in the manufacturing industry chain, electronic information industry chain and financial industry chain from the real world, constructed a complex network dataset, and evaluated our proposed model on the dataset. Specifically, we collected a large number of real records between companies, including upstream and downstream enterprise information, outbound investment, assets, profitability, personnel size, corporate geographic location, etc., as node characteristics. The topology structure of the network is shown in Figure. \ref{re}, and the link represents the cooperative relationship between nodes. Enterprises in the industrial chain not only form a horizontal cooperative relationship with enterprises of the same level, but also the production link of each intermediate product often involves vertical cooperation between multiple upstream suppliers and downstream retailers. The specific data set information is shown in Table .\ref{table-data}. We repeat the experiment 10 times with different random partitions and report the results based on the test accuracy of these 10 runs. We implemented the model in PyTorch and did all the training and testing tasks on two NVIDIA A100 GPUs.

\begin{table}[h]
\renewcommand{\arraystretch}{1.2}

\centering
\setlength{\tabcolsep}{3mm}
\begin{tabular}{c|ccc}
\hline
Items& Manufacturing & Electronic information & Financial\\
\hline

Nodes & 960 & 700 & 1500\\

Node features &32 & 32 & 16\\
Edges(thousand) & 250 & 160 & 600\\

Time span & 30 & 30 & 20\\

\hline
\end{tabular}
\caption{Detailed statistics of Datasets} 
\label{table-data}
\end{table}


\subsubsection{Topological structure prediction}

Topological structure prediction, that is, the part of neural network, our task is link prediction task, and five typical link prediction methods are selected as the baseline for experimental comparison:
\begin{itemize}
    \item \textbf{GBDT} \cite{friedman2001greedy} a classic tree-based ensemble learning predictor. GBDT is widely used in lots of prediction tasks among conventional financial institutions.
    \item \textbf{GCN} \cite{kipf2016semi} a traditional graph-based link prediction method. We first use graph convolution networks (GCNs) to encode the nodes in the training set to obtain the vector representation of the nodes, and then use these vector representations to perform supervised learning on the positive and negative samples in the training set (negative samples are resampled in each round of training).
    \item \textbf{TiRGN} \cite{li2022tirgn} a recent representation learning model for temporal knowledge graph reasoning that simultaneously considers sequential, repetitive, and cyclic patterns of historical facts in knowledge graphs. We follow its setting in the relationship prediction task.
    \item \textbf{STGNN} \cite{yang2021financial} an innovative financial risk analysis framework with graph-based supply chain mining. It employed a plain spatial-aware aggregator and temporal-aware aggregator for aggregating both neighbor and relational information in spatiotemporal Knowledge Graphs (STKG). 
    \item \textbf{JRCL} \cite{li2023learning} a novel framework to learn joint relational co-evolution in spatiotemporal Knowledge Graphs (STKG) for SMEs Supply Chain Prediction. It builds a large-scale financial STKG, which combines multi-view relationship sequence mining, relationship co-evolution learning and multiple random subspace layers to achieve the purpose of predicting the supply chain.

\end{itemize}

\subsubsection{Node state prediction}

Node state prediction, that is, the symbolic learning part. Our task is to predict the regression task of the future node state. We choose two classical time series prediction models and two physical information machine learning models as the baseline:

\begin{itemize}
    \item \textbf{LSTM} \cite{hochreiter1997long} a traditional model for processing time series. 
    \item \textbf{Transformer} \cite{vaswani2017attention} an effective method for time series forecasting. It uses the original transformer structure and does not introduce physical inspiration compared to our model.
    \item \textbf{PINN} \cite{raissi2018deep} a traditional supervised physical information machine learning model, which embeds differential equations as loss functions in neural networks, uses fully connected neural networks to fit the solutions of differential equations.
    \item \textbf{PhyCRNet} \cite{ren2022phycrnet} a physical-informed machine learning method for solving spatiotemporal partial differential equations (PDEs). Its loss function is defined as PDEs residuals, using auto-regression (AR), residual connection and ConvLSTM to learn temporal characteristics.
    
\end{itemize}

\subsection{Results and analysis}

\subsubsection{Topological structure prediction}\label{SRM}

In this part, we use two metrics to evaluate the performance of the model: accuracy (Acc) and F1-score (F1). The model performs well when both indicators obtain high values.

\begin{eqnarray}\label{accra} 
Acc&=&\frac{TP+TN}{TP+TN+FP+FN} \\
F1&=&2\times \frac{precision\times recall}{precision+recall}\\
precision &=& \frac{TP}{TP+FP}\\
recall &=& \frac{TP}{TP+FN}
\end{eqnarray}
where $TP, TN, FP$ and $FN$ represent true positive (positive class predicted as positive class), true negative (negative class predicted as negative class), false positive (negative class predicted as positive class), and true positive (positive class predicted as negative class), respectively.

Table. \ref{table_mining} has summarized the experimental results of performance comparison. Generally, our model can outperform the five selected typical baselines in various metrics, which also proves the effectiveness of our method.

\begin{table*}[h]

\renewcommand{\arraystretch}{1.5}
\setlength{\tabcolsep}{1.2mm}


\scriptsize
\scalebox{1}{
\begin{tabularx}{\textwidth}{c|cc|cc|cc}
\hline
&\multicolumn{2}{c}{Manufacturing} &\multicolumn{2}{c}{Electronic information} &\multicolumn{2}{c}{Financial}\\
\hline
Model & Acc $\uparrow$ &F1 $\uparrow$  & Acc $\uparrow$ &F1 $\uparrow$ & Acc $\uparrow$ &F1 $\uparrow$\\
\hline
GBDT & 0.761 $\pm$ 0.014 &0.784 $\pm $ 0.011 & 0.701 $\pm$ 0.024 &0.717 $\pm $ 0.015 &0.613 $\pm$ 0.033 &0.584 $\pm $ 0.031 \\

GCN&0.774 $\pm $ 0.013 &0.787 $\pm $ 0.027 &0.728 $\pm $ 0.013 &0.741 $\pm $ 0.017&0.671 $\pm $ 0.031 &0.668 $\pm $ 0.017 \\

TiRGN&0.807 $\pm $ 0.022 &0.800 $\pm $ 0.008&0.796 $\pm $ 0.022 &0.808 $\pm $ 0.016&0.735 $\pm $ 0.012 &0.780 $\pm $ 0.028\\

STGNN&0.887 $\pm$  0.014 &0.885 $\pm$ 0.011&0.886 $\pm$  0.011 &0.891 $\pm$ 0.015&\textbf{0.924 $\pm$  0.014} &0.929 $\pm$ 0.009 \\

JRCL & 0.898 $\pm$ 0.016 & 0.926 $\pm$ 0.018 & 0.891 $\pm$ 0.026 & 0.901 $\pm$ 0.015& 0.912 $\pm$ 0.021 & 0.926 $\pm$ 0.016\\

\textbf{Ours} &\textbf{0.907 $\pm $ 0.014 }&\textbf{0.942 $\pm $ 0.007} &\textbf{0.918 $\pm $ 0.014} &\textbf{0.907 $\pm $ 0.007}&0.920 $\pm $ 0.014 &\textbf{0.932 $\pm $ 0.007}

\\

\hline

\end{tabularx}}

\caption{Performance comparison of topological structure prediction.(The best result is in bold face.)}
\label{table_mining}
\end{table*}

Specifically, first of all, as a simple predictor, when only using some basic information of the enterprise as feature input, the performance of GBDT cannot achieve good results in various indicators, which shows that when dealing with industrial chain tasks, it is necessary to introduce graph neural networks as well as relationships between nodes to better handle this link prediction task for relational data. 
Secondly, GCN is used as a more original link prediction method. Although it uses a graph neural networks and considers the connection relationship between nodes, its performance is slightly better than that of ordinary GBDT. However, due to ignoring the time-varying characteristics of the information in the graph neural networks, its performance is still not very good.
Finally, compared with those recommendation methods based on static graphs, the temporal knowledge method TiRGN achieves further performance improvement by comprehensively considering the sequential independence of historical facts in the graph along the timeline. In addition, the spatiotemporal graph neural networks model STGNN also shows feasible performance by aggregating neighbor information and relationship information in the spatiotemporal graph neural networks. However, since only the spatial and temporal characteristics of nodes are taken into account, the practical significance of node links is not taken into account. For example, two companies are geographically close to each other, but one of them belongs to the steel manufacturing industry and the other belongs to the dairy product retail industry, so the probability of establishing a direct supply link is very small. Therefore, STGNN has not achieved optimal performance in this task.
It is worth noting that JRCL, as the most advanced method, has achieved a very similar effect to our method, but JRCL uses a more complex spatiotemporal network. There is a gap between the computational efficiency and our method.

In general, our model captures the spatiotemporal information of complex networks and can achieve remarkable performance in topological structure prediction tasks.

\subsubsection{Node state prediction}

In this section, in order to evaluate the accuracy of the solution generated by the model, we use two standard evaluation indicators, namely mean absolute error (MAE) and root mean square error (RMSE). The experimental results are shown in Table. \ref{table-develop}

\begin{eqnarray}
MAE &=& \frac{1}{N}\sum_{i=1}^{N}||\mathbf{\hat{u}}_i-\mathbf{u} _i||\\
RMSE &=& \frac{1}{N}\sum_{i=1}^{N}\sqrt{(|| \mathbf{\hat{u}}_i-\mathbf{u} _i || ^2_2)  }
\end{eqnarray}
where $\mathbf{u} _i$ and $\mathbf{\hat{u}}_i$ are the reference solution and the predicted solution. 

\begin{table}[h]
\renewcommand{\arraystretch}{1.2}

\centering
\setlength{\tabcolsep}{1mm}
\scriptsize
\scalebox{1}{
\begin{tabular}{c|cc|cc|cc}
\hline
&\multicolumn{2}{c}{Manufacturing} &\multicolumn{2}{c}{Electronic information} &\multicolumn{2}{c}{Financial}\\
\hline
 Model& MAE $\downarrow$ & RMSE $\downarrow$ & MAE $\downarrow$ & RMSE $\downarrow$ & MAE $\downarrow$ & RMSE $\downarrow$\\
\hline

LSTM & 0.327 $\pm $ 0.004& 0.419$\pm $0.010 & 0.316 $\pm $ 0.008& 0.419$\pm $0.012 & 0.385 $\pm $ 0.007& 0.484$\pm $0.007 \\
Transformer& 0.316 $ \pm $ 0.009 & 0.362 $\pm $ 0.014 & 0.308 $ \pm $ 0.012 & 0.362 $\pm $ 0.016 & 0.401 $ \pm $ 0.005 & 0.531 $\pm $ 0.012\\
PINN & 0.312 $\pm $ 0.014& 0.333$\pm $0.008 & 0.296 $\pm $ 0.012& 0.345$\pm $0.012 & 0.335 $\pm $ 0.015& 0.490$\pm $0.010 \\
PhyCRNet& 0.298 $\pm $ 0.016 & 0.397 $\pm $ 0.008 & 0.288 $\pm $ 0.015 & 0.339$\pm $ 0.020 & 0.366 $\pm $ 0.016 & 0.468 $\pm $ 0.018\\

\textbf{Ours} &\textbf{0.272 $\pm $ 0.007} &\textbf{0.351 $\pm $ 0.004} &\textbf{0.265 $\pm $ 0.010} &\textbf{0.338 $\pm $ 0.006} &\textbf{0.326 $\pm $ 0.007} &\textbf{0.429 $\pm $ 0.002}\\

\hline
\end{tabular}}

\caption{Performance comparison of node status prediction} 
\label{table-develop}
\end{table}

We experiment with the resulting differential equation as a loss function of PhyCRNet. PhyCRNet only needs initial conditions as input, and then performs unsupervised learning through autoregression and residual connection. In our experiment, it is equivalent to adding physical equation constraints to LSTM, so it can achieve better results. LSTM and Transformer, as traditional time series forecasting methods, are completely data-driven, so the effect is not good, but their advantage is that the calculation speed is faster. In addition, the five methods achieved similar results in manufacturing and electronic information data sets, but the opposite was true in financial data sets. Because the time interval of the financial industry data set is different, the advantage of using ODESolver is that we can solve the solution at any time.

\subsubsection{Resilience Evaluation}

We apply different degrees of perturbation to the initial moment, that is, randomly remove a different number of nodes, and for experimental convenience, we still keep the nodes, but change the node characteristics to 0, and disconnect the other nodes. A series of different times are selected to simulate the dynamic change of node state, and the node state curve is obtained.

\begin{figure}[h]
\centering
\subfloat[5\% attack]{\includegraphics[width=.49\textwidth]{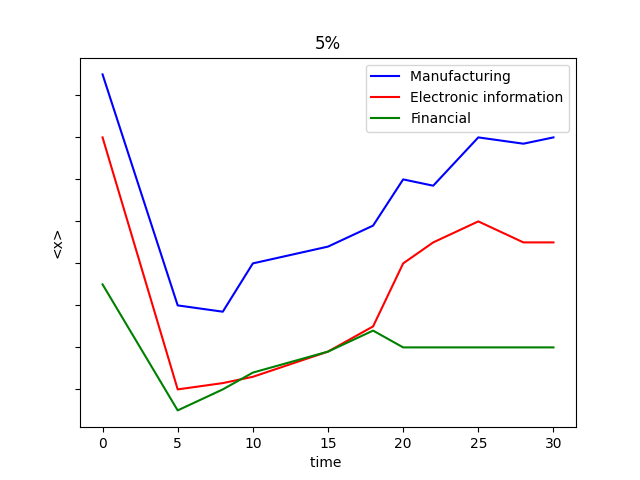}}
\label{5}
\hfill
\subfloat[10\% attack]{\includegraphics[width=.49\textwidth]{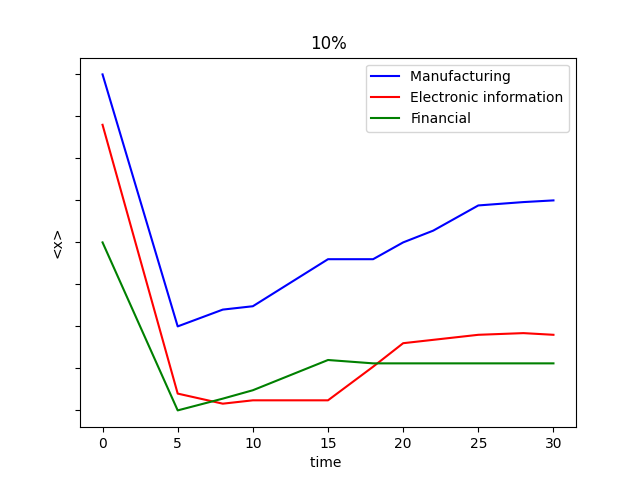}}
\label{10}
\hfill
\subfloat[20\% attack]{\includegraphics[width=.49\textwidth]{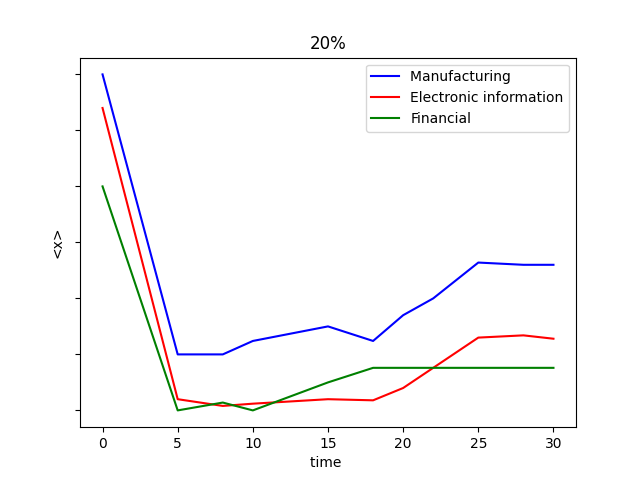}}
\label{20}
\hfill
\subfloat[50\% attack]{\includegraphics[width=.49\textwidth]{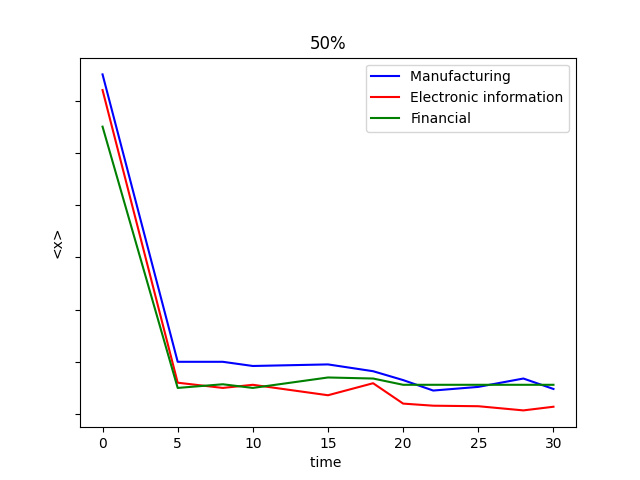}}
\caption{Node state curve facing different attacks}
\label{reli}
\end{figure}

The experimental results show that after applying a small attack (5\%, 10\%), the node state gradually deteriorates due to the existence of small damage and resilience, and after a period of time, the development trend will gradually recover and become stable, which is consistent with the description in Figure 3. Increased attack to 20\%. At this time, close to the critical value of seepage in complex networks, the node state will gradually recover, but the recovery speed is slow, and the network is still resilient. If the attacks further increase to 50\%, the main body of the network has been damaged, the network has a cascading failure, the development trend will become worse and worse, and finally the network crashes. In this case, the network is not resilient. In addition, through the experiment of three data sets, it can be seen that the larger the normal operation of the network, the stronger the anti-interference, the faster the recovery speed, that is, the stronger the resilience.

\section{Conclusion}

In this paper, aiming at the problems of insufficient accuracy and poor inter-pretability in complex network resilience prediction, we propose a physical information neural symbolic learning method for industrial chain resilience prediction. By using the co-evolution spatiotemporal graph network and the nonlinear ordinary differential equation, the dynamic model of large-scale physical entity activity state is established to predict the node state and link state in the future time respectively, and they are co-learned to predict the resilience of the industrial chain. The experimental results of different industrial chain data sets show that the proposed method can effectively and accurately predict the node state, and at the same time, due to the introduction of symbolic rules, it can still maintain a certain accuracy in the case of low data volume.

In the future work, we will explore the construction of more complex network relationships, such as the study of hypergraph network. In addition, the existing research is based on the known data in the past, and the real data is updated in real time. Take the industrial chain as an example, the industrial chain network will add new edges or new links to the old chain, and the network structure should be constantly updated. While traditional map neural network methods often show a significant decline in the performance of past tasks when learning new tasks, we will study more effective ways to achieve real-time monitoring and regulation of network resilience in the future


\acks{The authors would like to express their sincere gratitude to the anonymous reviewers for their insightful comments and constructive feedback, which greatly contributed to improving the overall quality of this paper. This work was supported in part by the National Key Research and Development Program of China under Grant 2022YFF0903300; and in part by National Natural Science Foundation of China under Grant 92167109}


\vskip 0.2in
\bibliography{sample}

\begin{thebibliography}{32}
\providecommand{\natexlab}[1]{#1}
\providecommand{\url}[1]{\texttt{#1}}
\expandafter\ifx\csname urlstyle\endcsname\relax
  \providecommand{\doi}[1]{doi: #1}\else
  \providecommand{\doi}{doi: \begingroup \urlstyle{rm}\Url}\fi

\bibitem[Artime et~al.(2024)Artime, Grassia, De~Domenico, Gleeson, Makse, Mangioni, Perc, and Radicchi]{artime2024robustness}
Oriol Artime, Marco Grassia, Manlio De~Domenico, James~P Gleeson, Hern{\'a}n~A Makse, Giuseppe Mangioni, Matja{\v{z}} Perc, and Filippo Radicchi.
\newblock Robustness and resilience of complex networks.
\newblock \emph{Nature Reviews Physics}, pages 1--18, 2024.

\bibitem[Feng et~al.(2024)Feng, Xu, Li, and Gao]{2024Analysis}
Xinghua Feng, Meihai Xu, Jianxin Li, and Ziyuan Gao.
\newblock Analysis of china's industrial network structure and its resilience from the sectoral perspective.
\newblock \emph{Habitat International}, 153, 2024.

\bibitem[Friedman(2001)]{friedman2001greedy}
Jerome~H Friedman.
\newblock Greedy function approximation: a gradient boosting machine.
\newblock \emph{Annals of statistics}, pages 1189--1232, 2001.

\bibitem[Gao et~al.(2016)Gao, Barzel, and Barab{\'a}si]{gao2016universal}
Jianxi Gao, Baruch Barzel, and Albert-L{\'a}szl{\'o} Barab{\'a}si.
\newblock Universal resilience patterns in complex networks.
\newblock \emph{Nature}, 530\penalty0 (7590):\penalty0 307--312, 2016.

\bibitem[Greydanus et~al.(2019)Greydanus, Dzamba, and Yosinski]{greydanus2019hamiltonian}
Samuel Greydanus, Misko Dzamba, and Jason Yosinski.
\newblock Hamiltonian neural networks.
\newblock \emph{Advances in neural information processing systems}, 32, 2019.

\bibitem[Hochreiter and Schmidhuber(1997)]{hochreiter1997long}
Sepp Hochreiter and J{\"u}rgen Schmidhuber.
\newblock Long short-term memory.
\newblock \emph{Neural computation}, 9\penalty0 (8):\penalty0 1735--1780, 1997.

\bibitem[Huang et~al.(2023)Huang, Sun, and Wang]{huang2023generalizing}
Zijie Huang, Yizhou Sun, and Wei Wang.
\newblock Generalizing graph ode for learning complex system dynamics across environments.
\newblock In \emph{Proceedings of the 29th ACM SIGKDD Conference on Knowledge Discovery and Data Mining}, pages 798--809, 2023.

\bibitem[Kipf and Welling(2016)]{kipf2016semi}
Thomas~N Kipf and Max Welling.
\newblock Semi-supervised classification with graph convolutional networks.
\newblock In \emph{International Conference on Learning Representations}, 2016.

\bibitem[Laurence et~al.(2019)Laurence, Doyon, Dub{\'e}, and Desrosiers]{laurence2019spectral}
Edward Laurence, Nicolas Doyon, Louis~J Dub{\'e}, and Patrick Desrosiers.
\newblock Spectral dimension reduction of complex dynamical networks.
\newblock \emph{Physical Review X}, 9\penalty0 (1):\penalty0 011042, 2019.

\bibitem[Li et~al.(2020)Li, Huang, Hong, Chen, Wu, and Zhu]{li2020closed}
Qing Li, Siyuan Huang, Yining Hong, Yixin Chen, Ying~Nian Wu, and Song-Chun Zhu.
\newblock Closed loop neural-symbolic learning via integrating neural perception, grammar parsing, and symbolic reasoning.
\newblock In \emph{International Conference on Machine Learning}, pages 5884--5894. PMLR, 2020.

\bibitem[Li et~al.(2024{\natexlab{a}})Li, Wang, Piao, Liao, and Li]{li2024predicting}
Ruikun Li, Huandong Wang, Jinghua Piao, Qingmin Liao, and Yong Li.
\newblock Predicting long-term dynamics of complex networks via identifying skeleton in hyperbolic space.
\newblock In \emph{Proceedings of the 30th ACM SIGKDD Conference on Knowledge Discovery and Data Mining}, pages 1655--1666, 2024{\natexlab{a}}.

\bibitem[Li et~al.(2024{\natexlab{b}})Li, Wang, Liu, and Shi]{li2024generalized}
Yibo Li, Xiao Wang, Hongrui Liu, and Chuan Shi.
\newblock A generalized neural diffusion framework on graphs.
\newblock In \emph{Proceedings of the AAAI Conference on Artificial Intelligence}, volume~38, pages 8707--8715, 2024{\natexlab{b}}.

\bibitem[Li et~al.(2023)Li, Zhu, Guo, Chen, Wang, Wang, Han, and Zhao]{li2023learning}
Youru Li, Zhenfeng Zhu, Xiaobo Guo, Linxun Chen, Zhouyin Wang, Yinmeng Wang, Bing Han, and Yao Zhao.
\newblock Learning joint relational co-evolution in spatial-temporal knowledge graph for smes supply chain prediction.
\newblock In \emph{Proceedings of the 29th ACM SIGKDD Conference on Knowledge Discovery and Data Mining}, pages 4426--4436, 2023.

\bibitem[Li et~al.(2022)Li, Sun, and Zhao]{li2022tirgn}
Yujia Li, Shiliang Sun, and Jing Zhao.
\newblock Tirgn: time-guided recurrent graph network with local-global historical patterns for temporal knowledge graph reasoning.
\newblock In \emph{Proceedings of the Thirty-First International Joint Conference on Artificial Intelligence, IJCAI 2022, Vienna, Austria, 23-29 July 2022}, pages 2152--2158. ijcai. org, 2022.

\bibitem[Liu et~al.(2024{\natexlab{a}})Liu, Ding, Song, and Li]{liu2024tdnetgen}
Chang Liu, Jingtao Ding, Yiwen Song, and Yong Li.
\newblock Tdnetgen: Empowering complex network resilience prediction with generative augmentation of topology and dynamics.
\newblock In \emph{Proceedings of the 30th ACM SIGKDD Conference on Knowledge Discovery and Data Mining}, pages 1875--1886, 2024{\natexlab{a}}.

\bibitem[Liu et~al.(2024{\natexlab{b}})Liu, Xu, Gao, Wang, Li, and Gao]{liu2024deep}
Chang Liu, Fengli Xu, Chen Gao, Zhaocheng Wang, Yong Li, and Jianxi Gao.
\newblock Deep learning resilience inference for complex networked systems.
\newblock \emph{Nature Communications}, 15\penalty0 (1):\penalty0 9203, 2024{\natexlab{b}}.

\bibitem[Ma et~al.(2023)Ma, Li, and Pan]{2023Global}
Li~Ma, Xiumin Li, and Yu~Pan.
\newblock Global industrial chain resilience research: Theory and measurement.
\newblock \emph{Systems}, 11\penalty0 (9), 2023.

\bibitem[Manhaeve et~al.(2018)Manhaeve, Dumancic, Kimmig, Demeester, and De~Raedt]{manhaeve2018deepproblog}
Robin Manhaeve, Sebastijan Dumancic, Angelika Kimmig, Thomas Demeester, and Luc De~Raedt.
\newblock Deepproblog: Neural probabilistic logic programming.
\newblock \emph{Advances in neural information processing systems}, 31, 2018.

\bibitem[Morone et~al.(2019)Morone, Del~Ferraro, and Makse]{morone2019k}
Flaviano Morone, Gino Del~Ferraro, and Hern{\'a}n~A Makse.
\newblock The k-core as a predictor of structural collapse in mutualistic ecosystems.
\newblock \emph{Nature physics}, 15\penalty0 (1):\penalty0 95--102, 2019.

\bibitem[Murphy et~al.(2021)Murphy, Laurence, and Allard]{murphy2021deep}
Charles Murphy, Edward Laurence, and Antoine Allard.
\newblock Deep learning of contagion dynamics on complex networks.
\newblock \emph{Nature Communications}, 12\penalty0 (1):\penalty0 4720, 2021.

\bibitem[Qu and Tang(2019)]{qu2019probabilistic}
Meng Qu and Jian Tang.
\newblock Probabilistic logic neural networks for reasoning.
\newblock \emph{Advances in neural information processing systems}, 32, 2019.

\bibitem[Raissi(2018)]{raissi2018deep}
Maziar Raissi.
\newblock Deep hidden physics models: Deep learning of nonlinear partial differential equations.
\newblock \emph{The Journal of Machine Learning Research}, 19\penalty0 (1):\penalty0 932--955, 2018.

\bibitem[Ren et~al.(2022)Ren, Rao, Liu, Wang, and Sun]{ren2022phycrnet}
Pu~Ren, Chengping Rao, Yang Liu, Jian-Xun Wang, and Hao Sun.
\newblock Phycrnet: Physics-informed convolutional-recurrent network for solving spatiotemporal pdes.
\newblock \emph{Computer Methods in Applied Mechanics and Engineering}, 389:\penalty0 114399, 2022.

\bibitem[Scheffer et~al.(2001)Scheffer, Carpenter, Foley, Folke, and Walker]{2001Catastrophic}
M.~Scheffer, S.~Carpenter, J.~A. Foley, C.~Folke, and B.~Walker.
\newblock Catastrophic shifts in ecosystems.
\newblock \emph{Nature}, 413:\penalty0 591--596, 2001.

\bibitem[Van~Steenkiste et~al.(2018)Van~Steenkiste, Chang, Greff, and Schmidhuber]{van2018relational}
Sjoerd Van~Steenkiste, Michael Chang, Klaus Greff, and J{\"u}rgen Schmidhuber.
\newblock Relational neural expectation maximization: Unsupervised discovery of objects and their interactions.
\newblock \emph{arXiv preprint arXiv:1802.10353}, 2018.

\bibitem[Vaswani et~al.(2017)Vaswani, Shazeer, Parmar, Uszkoreit, Jones, Gomez, Kaiser, and Polosukhin]{vaswani2017attention}
Ashish Vaswani, Noam Shazeer, Niki Parmar, Jakob Uszkoreit, Llion Jones, Aidan~N Gomez, {\L}ukasz Kaiser, and Illia Polosukhin.
\newblock Attention is all you need.
\newblock \emph{Advances in neural information processing systems}, 30, 2017.

\bibitem[Xie et~al.(2019)Xie, Xu, Kankanhalli, Meel, and Soh]{xie2019embedding}
Yaqi Xie, Ziwei Xu, Mohan~S Kankanhalli, Kuldeep~S Meel, and Harold Soh.
\newblock Embedding symbolic knowledge into deep networks.
\newblock \emph{Advances in neural information processing systems}, 32, 2019.

\bibitem[Yang et~al.(2021)Yang, Zhang, Zhou, Wang, Sun, Zhong, Fang, Yu, and Qi]{yang2021financial}
Shuo Yang, Zhiqiang Zhang, Jun Zhou, Yang Wang, Wang Sun, Xingyu Zhong, Yanming Fang, Quan Yu, and Yuan Qi.
\newblock Financial risk analysis for smes with graph-based supply chain mining.
\newblock In \emph{Proceedings of the Twenty-Ninth International Conference on International Joint Conferences on Artificial Intelligence}, pages 4661--4667, 2021.

\bibitem[Yu et~al.(2022)Yu, Yang, Wei, Li, and Pan]{yu2022probabilistic}
Dongran Yu, Bo~Yang, Qianhao Wei, Anchen Li, and Shirui Pan.
\newblock A probabilistic graphical model based on neural-symbolic reasoning for visual relationship detection.
\newblock In \emph{Proceedings of the IEEE/CVF Conference on Computer Vision and Pattern Recognition}, pages 10609--10618, 2022.

\bibitem[Yu et~al.(2023)Yu, Yang, Liu, Wang, and Pan]{yu2023survey}
Dongran Yu, Bo~Yang, Dayou Liu, Hui Wang, and Shirui Pan.
\newblock A survey on neural-symbolic learning systems.
\newblock \emph{Neural Networks}, 2023.

\bibitem[Zang and Wang(2020)]{zang2020neural}
Chengxi Zang and Fei Wang.
\newblock Neural dynamics on complex networks.
\newblock In \emph{Proceedings of the 26th ACM SIGKDD international conference on knowledge discovery \& data mining}, pages 892--902, 2020.

\bibitem[Zhang et~al.(2022)Zhang, Wang, Zhang, Havlin, and Gao]{zhang2022estimating}
Huixin Zhang, Qi~Wang, Weidong Zhang, Shlomo Havlin, and Jianxi Gao.
\newblock Estimating comparable distances to tipping points across mutualistic systems by scaled recovery rates.
\newblock \emph{Nature Ecology \& Evolution}, 6\penalty0 (10):\penalty0 1524--1536, 2022.

\end{thebibliography}

\end{document}